\documentclass[conference]{IEEEtran}
\IEEEoverridecommandlockouts
\usepackage[noadjust]{cite}
\usepackage{amsmath,amssymb,amsfonts}
\usepackage{algorithmic}
\usepackage{graphicx}
\usepackage{textcomp}
\usepackage{xcolor}
\usepackage{tabulary}
\def\BibTeX{{\rm B\kern-.05em{\sc i\kern-.025em b}\kern-.08em
    T\kern-.1667em\lower.7ex\hbox{E}\kern-.125emX}}
\begin{document}

\title{\Large \textbf{On the Effectiveness of Minisum Approval Voting in an Open Strategy Setting: An Agent-Based Approach}\\
}

\author{\IEEEauthorblockN{Joop van de Heijning}
\IEEEauthorblockA{\textit{Digital Age Research Center} \\
\textit{University of Klagenfurt}\\
Klagenfurt, Austria \\
johannes.vandeheijning@aau.at \\
ORCID 0000-0002-6246-9379}
\and
\IEEEauthorblockN{Stephan Leitner}
\IEEEauthorblockA{\textit{Department of Management Control} \\  \textit{and Strategic Management} \\
\textit{University of Klagenfurt}\\
Klagenfurt, Austria \\
stephan.leitner@aau.at \\
ORCID 0000-0001-6790-4651}
\and
\IEEEauthorblockN{Alexandra Rausch}
\IEEEauthorblockA{\textit{Department of Management Control} \\  \textit{and Strategic Management} \\
\textit{University of Klagenfurt}\\
Klagenfurt, Austria \\
alexandra.rausch@aau.at \\
ORCID 0000-0002-9275-252X}
}

\maketitle

\begin{abstract}
This work researches the impact of including a wider range of participants in the strategy-making process on the performance of organizations, which operate in either moderately or highly complex environments. Agent-based simulation demonstrates that the increased number of ideas generated from larger and diverse crowds and subsequent preference aggregation lead to the rapid discovery of higher peaks in the organization's performance landscape. However, this is not the case when the expansion in the number of participants is small. The results confirm the most frequently mentioned benefit in the Open Strategy literature: the discovery of better-performing strategies.
\end{abstract}

\textbf{\textit{Keywords-Open Strategy; minisum approval voting; strategy as a practice; \textit{NK} model; agent-based modeling.}}

\section{Introduction}
This paper shows that aggregating ideas from a diverse pool of participants using a specific aggregation mechanism leads to the rapid discovery of high-performing strategies. Including a wider range of participants in strategy-making is in line with a recent approach to strategy development, called Open Strategy (OS): OS is inclusive vs. restricted to the organizational elite, transparent vs. intransparent, and enabled by social information systems vs. merely supported by traditional IT \cite{Tavakoli2015}. Due to advances in (social) technology, changing societal norms, and several benefits, the interest in Open Strategy (OS) is on the rise \cite{Whittington2011}\cite{Seidl2019}.

Benefits of OS identified in empirical and conceptual research are the generation of better-performing strategies, increased buy-in and commitment, increased employee motivation, and improvements in an organization's reputation \cite{Sailer2018}. Tapping into the knowledge and intuition of nontraditional participants in the strategy process such as external stakeholders and access to a broader range of ideas are mentioned as reasons for the generation of better-performing strategies. The theory of the \textit{Wisdom of the Crowd} poses similar reasoning \cite{Silverman2007}.

In general, there is a lack of experimental evidence on when and how these claims might materialize, marked by calls for more longitudinal studies \cite{Seidl2019}. The goal of this work is to address this lack of evidence by evaluating with which number of participants and under which level of environmental complexity OS's most frequently mentioned benefit, the generation of better-performing strategies, eventuates. The results may contribute to a normative understanding of the OS approach and help guide academics and decision-makers towards better OS design.

This research turns to computational experimentation with an agent-based model for the following reasons \cite{Leitner2015Simulation-basedOverview}:
\begin{enumerate}
\renewcommand{\labelenumi}{(\alph{enumi})}
    \item Using empirical methods, it is impossible to disentangle the effects caused by the considered independent variables from the effects of other influences (the environment, competitors, the market, etc.) based on obtainable data such as from surveys and experiments, especially for strategic decision making. Additionally, on this topic, it is almost impossible to carry out longitudinal studies.
    \item Formal models frequently employed in economics would not be mathematically tractable due to their complexity.
\end{enumerate}

Agent-based simulation, the form of the simulation method employed herein, fits the research's objective: The firm and its stakeholders that participate in strategy-making are the agents. They are heterogeneous, boundedly rational, and act autonomously in an explicit space of local interactions. As summarized in \cite{Leitner2015OnMechanisms}, these characteristics of the model define an agent-based model \cite{Epstein1999Agent-basedScience}. The individual search processes, individual decision-making, and individual learning result in the macrobehavior of the firm, incrementally finding better-performing strategies.

This work makes use of the strategy as a practice perspective. In contrast to the planning and process view on strategy-making, the focus of strategy as a practice is on how the participants in strategy development act and interact with each other and with the organization \cite{Whittington1996}. To make sure we model aspects relevant to a realistic OS setting, we turn to the strategy as a practice perspective adapted to OS \cite{Tavakoli2017}.

The paper continues with a review of the literature in Section \ref{sec:lit} and a description of the model in Section \ref{sec:model}, discusses the simulation setup and the results in Section \ref{sec:results}, and ends with the conclusion and future work in Section \ref{sec:conclusion}.

\section{Literature Review}\label{sec:lit}

When implementing OS, organizations might opt for including both internal and external participants, or for limiting the inclusion scope to internal stakeholders only. Malhotra, Majchrzak, and Niemiec \cite{Malhotra2017} employ an action research approach and describe OS as an online collaboration process in which mainly external stakeholders are involved. They identify the risks of contentious conflict and self-promotion. Careful risk-mitigating actions led to a successful conclusion of the process. The company Siemens opted for directing inclusion internally by opening up strategy-making to all employees \cite{Hutter2017}. The case study shows that different forms of participation, i.e., commenting, evaluating, or merely submitting ideas, generate divergent effects on employees' engagement with the company, with the first two making a positive contribution, but not the latter.

Other lines of research, e.g., \cite{Mack2017}, elaborate on the impact of the organization's characteristics on the implementation of OS. The double case study in \cite{Mack2017} illustrates how a centralized organization tends to limit inclusion and transparency vs. a more decentralized organization. 

Other research investigates OS across time, i.e, during different phases. Research suggests that during the exploration of new products or markets, OS is more attractive than during the later phases in a product cycle \cite{Appleyard2017}. In addition, \cite{Hautz2019PracticesStrategy} and \cite{Hautz2017a} highlight that also within the strategy process temporal effects exist: Organizations might do better by opening up strategy-making during the generation of ideas for new strategies than during the subsequent selection of new strategies.

There are social network connections among stakeholders and between the organization and its stakeholders. Four themes related to the intersection between network research and OS are identified in \cite{Hautz2019AStrategy}, leading to the following advantages of the network perspective when studying OS: First, it is possible to move toward a more relational understanding. Second, the availability of data in IT-based initiatives in OS provides opportunities to apply quantitative methods. Third, it could enable multilevel research. Finally, network analysis can be highly appropriate for studying the outcomes of increased openness.

While OS is associated with a wide range of beneficial outcomes, trade-offs exist \cite{Hautz2017}. E.g., the possibility of obtaining better-performing strategies is reported as a benefit in \cite{Malhotra2017}, while there is a risk of loss of commitment among participants when expectations about the impact of contributions remain unmet \cite{Baptista2017SocialStrategy}. Long term studies on how better-performing strategies emerge are, however, still missing.

\section{Model Description}\label{sec:model}
The purpose of the model is to investigate how the number of participants and the level of complexity of the environment affects an organization's performance in an OS approach.

The strategy-making process is simulated in the \textit{Open Strategy as a Practice} framework introduced in \cite{Tavakoli2017}. The three main components of the framework are a) the practitioners representing the people making strategy, b) the praxis component standing for what happens in an iterative process with the episodes taking place in a certain organizational context, and finally c) the set of practices representing the tools and mechanisms used to develop a strategy. When strategy-making is open, the set of practitioners includes stakeholders along with the firm's upper echelon, some types of voting mechanisms are typically part of the practices, and the praxis is transparent by including feedback to the stakeholders involved.

The model features a single firm that seeks and implements high-performing strategies in an iterative manner. The firm may or may not choose to include stakeholders in the strategy-making process. It operates in a static, complex environment.

The firm is represented by its upper management and it is always included in the simulation as $P_1$ in the set of practitioners $\{P_j: j\in\{1,..,S\}\}$ where $S$ is the total number of practitioners. In an OS setting, strategy-making also includes $S-1\geq1$ stakeholders from inside or outside the firm such as employees and customers. While this work studies the performance of strategies for the firm that are being discovered and implemented, all practitioners are assumed to gain a personal utility from these strategies and want to maximize their utility.

The praxis is modeled as a cyclic process in the order of months, where each episode $t\in\{1,..,T\}$ consists of phases as in \cite{Tavakoli2017}. Hautz \cite{Hautz2017a} and references therein propose that in a strategy-making process, first comes generating a range of strategy ideas, next comes selecting the most appropriate one, followed by implementation. Mack and Szulanski \cite{Mack2017} present case illustrations showing that a similar framework can describe reality. As in \cite{Sailer2018}, we distinguish the following four phases in an episode: 1) preparation phase, 2) generation phase, 3) selection phase, and 4) implementation phase. Strategies are modeled as bitstrings $\textbf{s}\in\{0,1\}^N$ consisting of $N$ binary decisions $s_i,\ i\in\{1,..,N\}$. Parameter $K$ signifies the number of interactions between decisions and can be considered as a proxy for how tightly departments within an organization are interwoven, thereby also shaping the complexity of the task environment  \cite{Levinthal1997}.

\begin{figure}[htbp]
\centerline{\includegraphics[width=\linewidth]{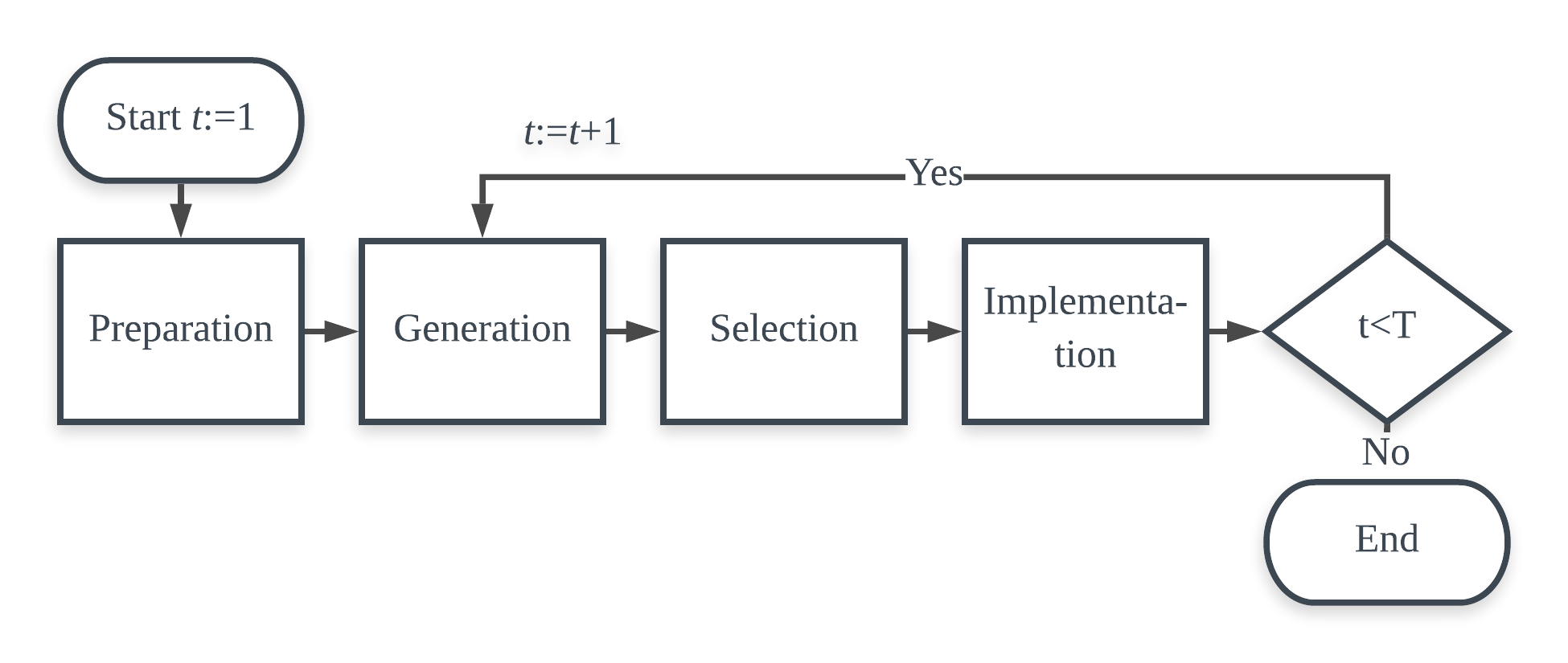}}
\caption{Flow diagram of one simulation.}
\label{fig:flow}
\end{figure}

In condensed form, the sequence of events is as follows (see Figure \ref{fig:flow}): The simulation runs for $T$ episodes with the number of stakeholders $S$ and the number of decision interactions $K$ as independent variables. In each episode, 1) the simulation's environment is set up in the preparation phase, 2) the practitioners come up with ideas for a new strategy, and a preference aggregation mechanism distills these ideas to a shortlist in the generation phase, 3) the practitioners vote for the new strategy in the selection phase, and 4) the performance for the firm of the new strategy is recorded as the dependent variable depending on independent variables $S$ and $K$, and choices made in the OS process such as the type of the aggregation mechanism used, in the implementation phase. The subsections below describe the four phases and the model in more detail.

\subsection{Preparation Phase}

The firm starts with a random current strategy $\mathbf{s}_{cur}$. The simulator (the simulation software is written in Python 3.7) also generates the practitioners' utilities of the strategies the firm might implement. Strategies are mapped to performances for practitioners $P_j$ in \textit{performance landscapes} $F_j:\{0,1\}^N\rightarrow[0,1]$ defined by the $NK$ model \cite{Kauffman1989}. I.e., the strategies related to higher numerical values correspond to high-performing variants in the space of possibilities, and vice versa. A higher $K\in\{0,N-1\}$ leads to rugged, highly nonlinear performance landscapes with more local peaks, whereas a lower $K$ gives rise to a smoother search space, see also \cite{Csaszar2018a}.

As all practitioners including the firm are assumed to be correlated in their preferences toward strategies, the simulation generates random performance landscapes $F_j(\mathbf{s})$ with a pairwise correlation coefficient $\rho$ in line with \cite{Verel2013} (with a random interaction matrix). To account for the diversity in the pairwise correlations, the algorithm in \cite{Verel2013} is extended to use a perturbed correlation matrix \cite{Marsaglia1984}.

Evaluating strategies is not without error: Assuming limited cognitive capacity \cite{Simon1955}, practitioners' views of their landscapes are somewhat obfuscated. Hence, every time a practitioner $P_j$ evaluates a strategy \textbf{s}, the model adds a random error term $\epsilon$ from a Gaussian with mean 0 and standard deviation $E_j$:
\begin{equation}
\begin{split}
    F_j(\mathbf{s})'&=F_j(\mathbf{s})+\epsilon,\\
    \epsilon &\sim \mathcal{N}(0,E_j).
\end{split}
\end{equation}
We assume practitioners are diverse in their cognitive capacities. Consequently, at the start of a simulation, we draw random variables $D_j,\  j\in\{1,..,S\}$ from a Gaussian with mean 0 and standard deviation $E$ and then take the absolute value of them to obtain the individualized $E_j$:
\begin{equation}
\begin{split}
    E_j&=|D_j|,\\
    D_j &\sim \mathcal{N}(0,E).
\end{split}
\end{equation}

This phase takes place in episode $t=1$ only, otherwise, the episode starts off with the generation phase directly.

\subsection{Generation Phase}

This phase generates a shortlist of $L$ \textit{candidates}; strategies that can be taken into consideration in the selection phase when practitioners vote for the firm's strategy in $t+1$. As the firm has a limited capacity for change in each episode, \textit{appropriate} candidates $\mathbf{s}$ for $t+1$ do not differ in more than $C$ decisions from $\mathbf{s}_{cur}$, i.e., they have a \textit{Hamming distance} $d_H(\mathbf{s}, \mathbf{s}_{cur})$ of $C$ decisions maximum, assuming $N>C$.

As the first step in this phase, each practitioner enters one idea from the set of appropriate candidates for the firm's new strategy in $t+1$ on a list of ideas. 
Entering ideas is modeled as follows: Practitioners observe the firm's current strategy $\mathbf{s}_{cur}$. Assuming limited cognitive capacity \cite{Simon1955}, each practitioner imagines a personal random subset of only $Q$ strategies out of the set of appropriate candidates. They rank these $Q$ strategies by evaluating them one by one in their performance landscapes. Every practitioner then enters their preferred strategy on the list of $S$ ideas. Duplicates may occur.

As the second and last step in this phase, the preference aggregation mechanism \textit{minisum approval voting} \cite{Brandt2016} distills these ideas into a shortlist of $L$ candidates for the selection phase in this episode $t$. Let $\mathbf{s}$ be an appropriate candidate for the firm's strategy in $t+1$ and let $\mathbf{s}_j$ be $P_j$'s idea. Then, the minisum score for $\mathbf{s}$ equals $\sum_{j=1}^S d_H(\mathbf{s},\mathbf{s}_j)$, the sum of the Hamming distances between $\mathbf{s}$ and all $\mathbf{s}_j$. The candidates are ranked by score (candidates with the same score are ranked in random order) and the $L$ lowest-ranked candidates win. As this algorithm takes the sum of distances, discontent from single practitioners with particular candidates may not influence the ranking. Therefore, minisum approval voting is a \textit{utilitarian} preference aggregation mechanism.

\subsection{Selection Phase}

The shortlist of $L$ candidates is extended by $\mathbf{s}_{cur}$. The practitioners evaluate the $L+1$ candidates and communicate their rankings to the \textit{Borda count} voting rule from the set of practices \cite{Brandt2016}. Every time a practitioner ranks a candidate first, the candidate adds $L$ scores to its total count. Every practitioner's second ranking rewards a candidate with $L-1$ scores, etc., with a last ranking giving zero scores. The winning candidate $\mathbf{s}_{new}$ is the candidate with the highest total count. If there are multiple candidates with the maximum total count, the simulation picks one at random.

\subsection{Implementation Phase}

The firm communicates $\mathbf{s}_{new}$ to its stakeholders and implements the strategy, thereby ending episode $t$. The model evaluates $\mathbf{s}_{new}$ in the firm's landscape and stores the result, relating time step $t$ to performance $F_1(\textbf{s}(t))$ as the dependent variable. If $t<T$, the simulation, skipping over the preparation phase, continues with the generation phase in $t+1$. The winning candidate $\mathbf{s}_{new}$ now has become $\mathbf{s}_{cur}$ in $t+1$.

\section{Results}\label{sec:results}

The previous section presented features of the model that are relevant when considering the impact of multiple practitioners on strategy-making. The subsections below define the simulation setup and discuss the results.

\subsection{Simulation Setup}
Table \ref{tab:variables} shows the variables used in the model. Starting with the control variables, we set $N=10$ giving sufficient combinatorial richness while not putting too much demand on computational resources. The pairwise correlation coefficient $\rho$ is set to the moderate positive value of 0.5. $C$, the maximum Hamming distance with the current strategy that makes a strategy a viable candidate, equals 2; while constraining the strategy updates in subsequent episodes to only 2 decisions, it still leaves plenty of choices for practitioners to imagine ideas as the set of appropriate candidates scales with the sum of Binomial coefficients $\binom{N}{k}$, with $k\in\{1,..,Q\}$. The number of ideas per practitioner $Q$ is set to 2 due to the assumption of limited cognitive capacity \cite{Simon1955}. The size $L$ of the shortlist of candidates for selection is set to 3 as to not overburden voters in the selection phase. With $E=\frac{1}{16}$, the evaluation error $E_j$ will average out to a small $\frac{E\sqrt{2}}{\sqrt{\pi}}\approx 0.05$ as the mean of the folded normal distribution with standard deviation $E$. The number of episodes $T$ in a simulation is set to 100 as in all cases the results stabilize at this value.
\begin{table*}[htbp]
\caption{Variables}
\begin{center}
\begin{tabulary}{\linewidth}{lllL}
\hline\noalign{\smallskip}
Classification & Symbol & Value/Range & Description  \\
\noalign{\smallskip}\hline\noalign{\smallskip}
Control     & $N$ & 10 & The number of binary decisions in a strategy \textbf{s}.       \\
            & $\rho$ &0.5& The average correlation coefficient between practitioners' performance landscapes.  \\
            & $C$ &2& Maximum Hamming distance from the current strategy that makes a strategy an appropriate candidate. \\
            & $Q$ &2& The number of strategies practitioners can imagine in the generation phase. \\
            & $L$ &3& The size of the shortlist of candidates for the next episode's strategy. \\
            & $E_j$ &0.05& Standard deviation of practitioner's $P_j$ evaluation error.\\ 
            & $E$ &0.0625& Standard deviation of each $E_j$. \\
            & $T$  &100& The number of episodes per simulation.   \\
Independent   
            & $K$ &$\{4,7\}$& The number of interactions between decisions. \\
            & $S$ &$\{1,10,100\}$& The number of practitioners. \\
Dependent 
            & $F_1(\textbf{s}(t))$  &$[0,1]$& Performance for the firm of strategy $\mathbf{s}$ at episode $t$.\\
            \noalign{\smallskip}\hline
\end{tabulary}
\label{tab:variables}
\end{center}
\end{table*}

The independent variables are $K$ and $S$. The number of decision interactions $K$ is varied to a moderate setting, i.e., 4 and a high setting, i.e., 7 for contrast. The number of practitioners $S$ equals either 1, when strategy-making is exclusive for the firm as in a traditional Closed Strategy setting, and 10 or 100 when stakeholders are introduced in an OS setting.

The dependent variable $F_1(\mathbf{s}(t))$, the performance for the firm at time $t$, is recorded at the end of every episode from $t=1$ to $T$ with $F_1$ normalized to $[0,1]$. To mitigate random effects, it was found sufficient to average the scenarios over 4000 simulation repetitions \cite{Lorscheid2012}. Confidence intervals over $F_1(\mathbf{s}(t))$ are calculated at 95\%.

\subsection{Simulation Results and Discussion}
\begin{figure}[htbp]
\centerline{\includegraphics[width=\linewidth]{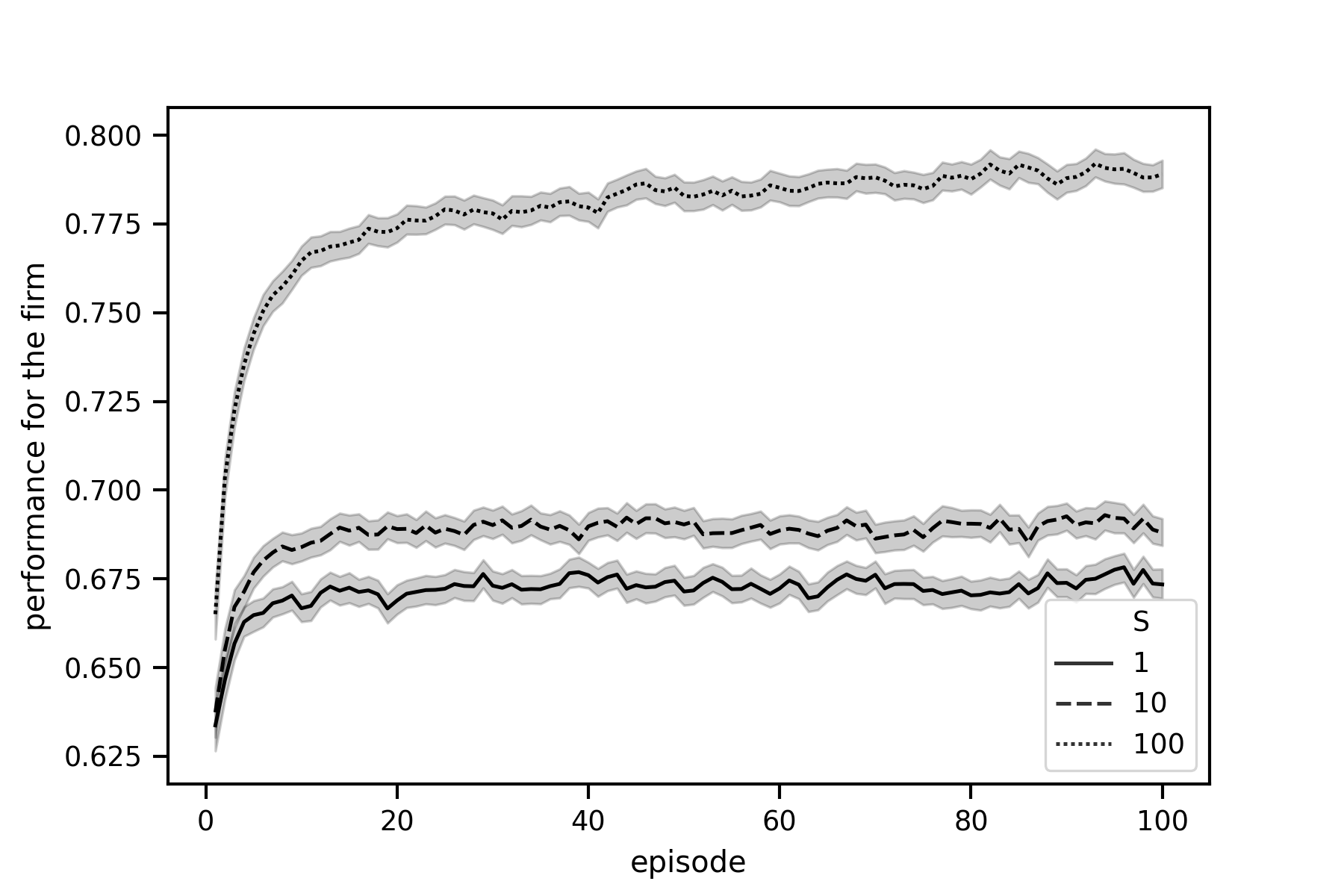}}
\caption{Graph of performance for the firm over episodes at $K=4$.}
\label{fig:K4}
\end{figure}

Figure \ref{fig:K4} shows the performance of the firm $F_1(\mathbf{s}(t))$ over one hundred episodes at the moderate level of interactions $K=4$. Included are results for $S=1$, when strategy-making is restricted to the firm as the only practitioner involved in strategy development. When $S=10$ or $S=100$, strategy-making is opened up to include an additional nine, respectively, 99 practitioners, which are stakeholders of the firm. In $t=1$, just one episode after starting with a random strategy, Figure \ref{fig:K4} shows already a significant difference between performance at $S=1$ or $S=10$ on the one hand and $S=100$ on the other. At $t=5$, also the OS setting with ten practitioners starts outperforming the closed strategy-making. Over the entire range of episodes, $S=100$ outperforms both $S=10$ and $S=1$.

\begin{figure}[htbp]
\centerline{\includegraphics[width=\linewidth]{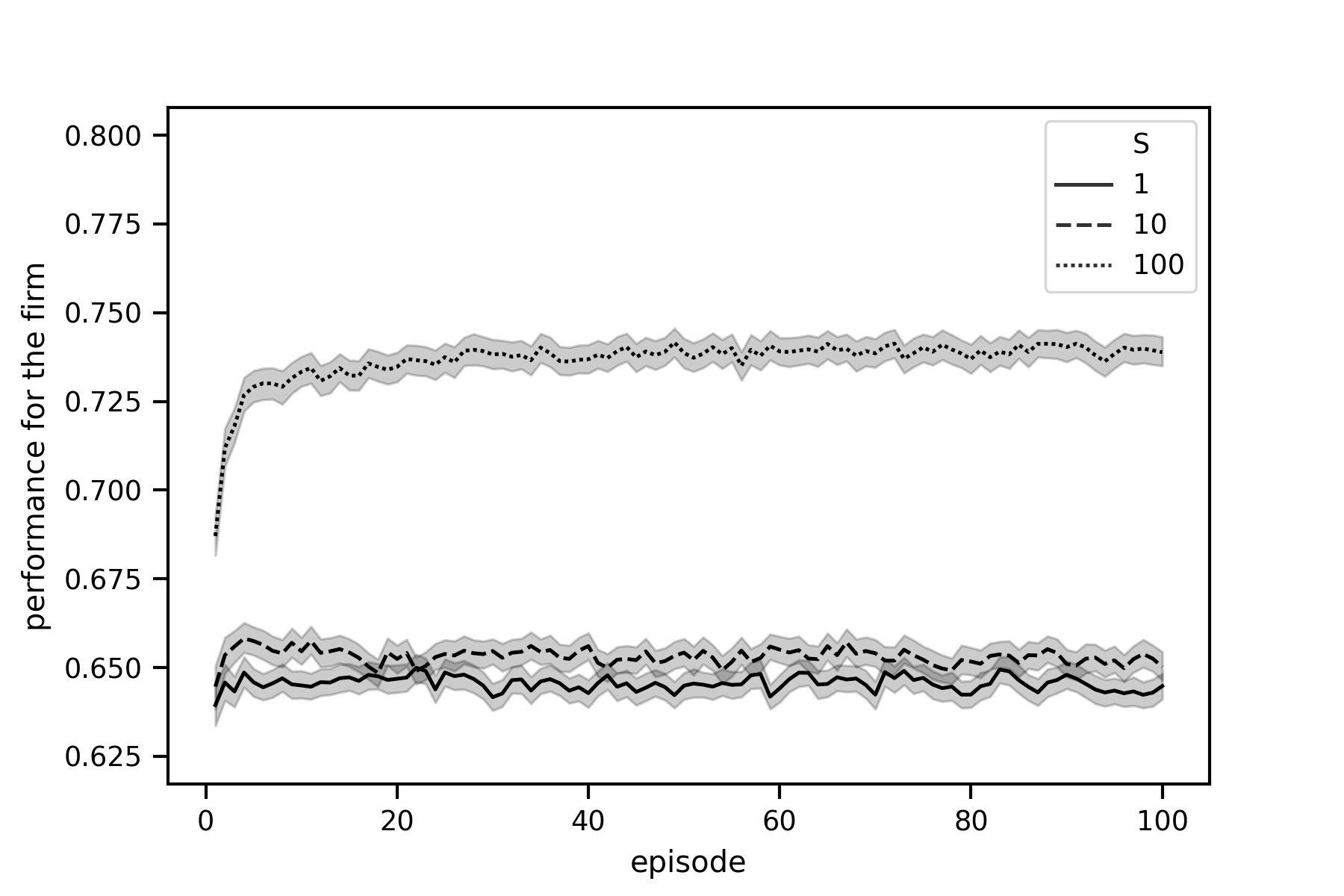}}
\caption{Graph of performance for the firm over episodes at $K=7$.}
\label{fig:K7}
\end{figure}

Figure \ref{fig:K7} illustrates that when $K=7$, the additional ruggedness of the landscapes produces lower performances for all values of $S$ as it is more difficult to find peaks with high performing strategies for the firm. Still, $S=100$ outperforms both $S=1$ and $S=10$. In contrast to $K=4$, with $K=7$, the OS setting with ten practitioners does not significantly outperform the Closed Strategy setting.

The results demonstrate that in our model the performance improves or is at least the same when opening up strategy-making for both $K=4$ and $K=7$, already in the initial episodes. This finding confirms the most frequently mentioned benefit of an OS approach: the discovery of better-performing strategies \cite{Sailer2018}. In a Closed Strategy setting, the firm has access to fewer ideas; access to a larger, diverse pool of ideas helps in exploring the landscape faster. In a sense, an OS approach gives a wider view, without needing to resort to uncertain big jumps in the searching process as in \cite{Levinthal1997}.

The result that with higher $K$ the advantage of an OS approach for $S=10$ seems to disappear can be explained as follows: The theory of the Wisdom of the Crowd states that in many discovery processes a group outperforms the individual, even if the single person is an expert \cite{Silverman2007}. Wisdom of the Crowd is especially effective when individual judgments cluster around the correct central value \cite{Lyon2013}. However, even when judgments among a wide group average to the central value, querying a small subset of the group can have a detrimental effect on the discovery of the correct value \cite{Csaszar2018}. In our model, while the practitioners' performance landscapes are correlated, there is nothing that guarantees that the "central value" in the set of landscapes is the firm's landscape. Our findings suggest that when a landscape is very rugged, a small bias can have large effects, countering the advantage of the added number and diversity of ideas that an OS approach can bring. Sensitivity analysis with variations in $\rho$ confirms the expectation that in ceteris paribus, a higher correlation among practitioners leads to higher performance.

\section{Conclusion and Future Work}\label{sec:conclusion}

The results suggest that the often stated benefit of an OS approach, namely, the discovery of better-performing strategies \cite{Sailer2018}, can indeed be obtained through preference aggregation of a diverse group of practitioners with minisum approval voting, even if voters' preferences are not 100\% aligned with the firm's preference. Moreover, the results indicate that a larger group of practitioners ($S=100$) outperforms a smaller group ($S=10)$ in organizations operating in both moderately ($K=4$) and highly ($K=7$) complex environments. With $K=7$, when it is more difficult to navigate the firm's landscape, the advantage of OS, at least with a smaller number of participants, disappears due to the drawback of practitioners' preferences that are on average not necessarily 100\% aligned with the firm's preference.

A limitation of our methodology is that external validation is hard, as data is difficult to obtain by empirical methods \cite{Davis2007DevelopingMethods}. To focus on the core aspects of the research question, much complexity, such as communication between stakeholders, the effects of memory, and network effects, was eliminated that might capture critical aspects of reality.  Additionally, this paper would benefit from sensitivity analysis over the control variables.

Future work can extend the model by including network effects among practitioners, strategic voting, and the consequences of voter dissatisfaction. Simulations could consider the effects of restricting the opening of strategy-making to specific phases in the strategy process or investigate the impact of temporal opening up depending on exploratory or exploitative demands in organizations' life cycles. Furthermore, egalitarian vs. utilitarian preference aggregation mechanisms \cite{Baumeister2015VoterElections} can be evaluated for their impact on the drawback of loss of commitment by dissatisfied practitioners \cite{Hautz2017}. Additional research can also investigate the apparent dilemma between the posed benefits of a broader range of ideas \cite{Chesbrough2007} (high diversity in preferences) and the requirement of low bias (low diversity) observed in this research.

\bibliographystyle{IEEEtran}
\bibliography{references.bib}

\end{document}